\documentclass[letterpaper, 10 pt, conference]{ieeeconf}  

\usepackage{algorithm, algorithmic}
\usepackage{booktabs} 
\usepackage{graphicx,graphics,amsthm,amssymb,enumerate,tikz}
\usepackage{mathtools}
\usepackage{arydshln}
\usetikzlibrary{positioning,calc}
\usepackage{fancyhdr}

\theoremstyle{plain}
\newtheorem{theorem}{Theorem}

\newtheorem{lemma}{Lemma}

\theoremstyle{definition}
\newtheorem{definition}{Definition}
\newtheorem{assumption}{Assumption}

\theoremstyle{remark}
\newtheorem{remark}{Remark}

\def\R{\mathbb{R}}
\def\z{\mathbf{z}}

\def\V{\mathcal{L} \, V}

\IEEEoverridecommandlockouts                              

\title{\LARGE \bf
Fully Distributed GNE Algorithms for Multi-Robot Placement \\without Consensus on Multipliers
}

\author{Shao-An Yin, Mingyi Hong, and Nicola Elia%
\thanks{* This work was supported in part by NSF ECPN award 2311007.}%
\thanks{All authors are with the Department of Electrical and Computer Engineering,
University of Minnesota, USA. {\small\texttt{yin00425@umn.edu, mhong@umn.edu, nelia@umn.edu}}.}}
\fancypagestyle{firstpage}{
    \fancyhf{}

    \fancyfoot[C]{\footnotesize
    \copyright~2026 IEEE. Personal use of this material is permitted.
    Permission from IEEE must be obtained for all other uses.}
}

\begin{document}

\maketitle
\thispagestyle{empty}
\pagestyle{empty}
\thispagestyle{firstpage}

\begin{abstract}
Recent machine learning research has increasingly focused on equilibrium analysis in non-cooperative games rather than solely on optimal solutions. Many such problems involve shared constraints and can be formulated as Generalized Nash Equilibrium Problems (GNEPs). For strongly monotone games, existing methods compute consensus-based variational GNEs (v-GNEs) by exchanging Lagrange multipliers. We propose a fully distributed continuous-time algorithm for shared linear equality constraints that converges without multiplier exchange and reaches any GNE, reducing communication overhead and improving privacy. Discrete-time schemes are also provided, and the method is validated on a multi-robot placement task.
\end{abstract}

\section{Introduction}
Nash Equilibrium (NE) has been widely studied across diverse areas \cite{NEURIPS2022_24f420aa} and plays a central role in multi-agent and adversarial machine learning, including multi-agent reinforcement learning \cite{NashQ, littman_1994, meanfield} and generative modeling frameworks \cite{GAN, pmlr-v119-jin20e}. In many settings, agents must compute equilibria under shared constraints \cite{NEURIPS2021_174a61b0, tsaknakis_minimax_2023, pmlr-v48-pedregosa16}, leading to the Generalized Nash Equilibrium Problem (GNEP).

The concept of GNEP was introduced in \cite{debreu-social-1952, rosen-existence-1965}. In this setting, each agent minimizes its local cost given others’ decisions, while its feasible set also depends on those decisions, often due to shared resources or safety constraints. A Generalized Nash Equilibrium (GNE) occurs when no agent can reduce its cost by unilaterally changing its decision within this feasible set. Early work focused on existence results \cite{debreu-social-1952, rosen-existence-1965}. However, the existence of a solution does not imply that it can be easily computed, even with centralized coordination.

Early centralized approaches, such as \cite{fukushima-restricted-2011}, solved GNEPs using augmented penalty methods that repeatedly compute Nash equilibria with increasing penalties on shared constraints, later extended in \cite{facchinei-penalty-2010, kanzow-augmented-2016, dreves-nonsmooth-2011}. However, the resulting nonsmooth inner problems make distributed implementations difficult \cite{dreves-nonsmooth-2011}. Moreover, centralized computation becomes impractical with many agents due to high computational and communication costs, and agents may be unwilling to share their cost functions for privacy reasons. These challenges motivate distributed approaches where agents compute decisions with limited information exchange.

Most distributed GNEP studies focus on variational GNEs (v-GNEs) for strongly monotone games with linear shared constraints. For example, \cite{yi-pavel-2019, cenedese-asynchronous-2021, huang-distributed-2021} use operator-splitting methods, while \cite{bianchi-continuous-time-2021} adopts a continuous-time approach. These methods rely on equilibrium uniqueness and enforce multiplier consensus, thus computing only v-GNEs. In a v-GNE, agents subject to the same shared constraints use identical multipliers, implicitly assuming a central authority that enforces feasibility and broadcasts a common shadow price. In contrast, general GNEs allow agent-specific multipliers, modeling personalized tariffs on shared resources. 


In this paper, we formulate the multi-robot placement problem as a distributed GNEP with individual convex constraints and shared linear equality constraints. For this class of problems, we propose a fully distributed continuous-time primal--dual algorithm that eliminates multiplier exchange and converges to general GNEs rather than only v-GNEs. We also establish convergence for strongly monotone games with shared linear equality constraints. The main contributions of this work are:
\begin{enumerate}
    \item To the best of our knowledge, this is the first distributed algorithm that avoids consensus on Lagrangian multipliers for general convex cost functions, thereby reducing per-iteration communication.
    \item Existing distributed algorithms are typically restricted to v-GNEs. In contrast, depending on the initialization, our algorithm can converge to general GNEs.
    \item We formulate the multi-robot placement problem with shared constraints as a GNEP and demonstrate the effectiveness of our algorithm. For this class of problems, our method reduces information exchange to one-third per iteration compared to existing approaches.
\end{enumerate}

\section{Preliminaries}
\subsection{Notations}
We denote $\mathbb{R}^m$ as the $m$-dimensional Euclidean space, $\R_{\ge 0}$ and $\R_{>0}$ as the nonnegative and positive reals, and $[m]=\{1,2,\dots,m\}$. Vectors $\mathbf{1}$ and $\mathbf{0}$ denote all-ones and all-zeros of appropriate dimensions. For $a \in \R^m$, $(a)_i$ is its $i$-th component, $a^\top$ its transpose, and $a^\top b$ the inner product with $b \in \R^m$; $a \le b$ indicates component-wise inequality. For $A \in \R^{p \times q}$, $(A)_{ij}$ is the $(i,j)$ entry and $A^\top$ the transpose. $\mathcal{G} = (\mathcal{V}, \mathcal{E})$ denotes a graph with vertex set $\mathcal{V}$ and edge set $\mathcal{E}$. Following~\cite{cherukuri-asymptotic-2016,ebrahimi-robust-2019}, we also define the projection operator:
\begin{equation}
    [a]^+_b := \begin{cases}
        a, & \text{if }b > 0 \\
        max \{ 0, a\} & \text{if }b = 0
    \end{cases}.
\end{equation}

Lastly, the sublevel set of a function $V: \R^m \to \R$ with $\beta>0$ is defined as $V^{-1}(\le \beta) = \{ x \in \R^m \, | \, V(x) \le \beta\}$.

\subsection{Generalized Nash Equilibrium Problems (GNEPs)}
Consider a GNEP with $N$ agents. Each agent $v$ controls decision variables $z_v \in \mathbb{R}^{n_v}$ and has a cost function $f_v(z_v, \z_{-v}^f)$, which depends on its own decision variables and those of other agents, where $\z_{-v}^f$ denotes the variables from agents that influence $f_v(\cdot)$. This dependency induces a cost-function graph $\mathcal{G}_f = (\mathcal{V}_f, \mathcal{E}_f)$. Each agent is also subject to individual constraints $h_v(z_v) \le \mathbf{0}_{q_v}$. Let $\z \in \mathbb{R}^n$, with $n = \sum_{v=1}^N n_v$, denote the joint decision vector of all agents, and let $q = \sum_{v=1}^N q_v$ denote the total number of individual constraints.

Additionally, the constraint set of each agent depends on the decisions of other agents. Specifically, $z_v$ must satisfy $g_v(z_v, \z_{-v}^g) = \mathbf{0}_{m_v}$, where $\z_{-v}^g$ denotes the collection of decision variables from all agents other than $v$ that influence $g_v(\cdot)$. This dependency induces a constraint graph $\mathcal{G}_g = (\mathcal{V}_g, \mathcal{E}_g)$. Formally, each agent $v$ solves the following optimization problem:
\begin{equation}\label{equ:GNEP}
    \begin{split}
        \begin{matrix*}[l]
            \min_{z_v} & f_v(z_v,\, \z_{-v}^f) \\
            \text{subject to} & h_v(z_v) \le \mathbf{0}_{q_v}\\
            & g_{v}(z_v,\, \z_{-v}^g) =  \mathbf{0}_{m_v}
        \end{matrix*}
    \end{split}.
\end{equation}
To avoid abuse of notation, we use $\z_{-v}$ to denote the collection of other agents' decision variables when the context is clear. That is, we write $f_v(z_v, \z_{-v})$ instead of $f_v(z_v, \z_{-v}^f)$ and $g_v(z_v, \z_{-v})$ instead of $g_v(z_v, \z_{-v}^g)$.

Specifically, in the paper, we assume the presence of globally shared linear equality constraints. In other words,  
\begin{equation}\label{equ:linear}
\begin{split}
    &g(\z) = A \, \z -\mathbf{c}  \\
\end{split},
\end{equation}
where $A \in \mathbb{R}^{p \times n}$ and $\mathbf{c} \in \mathbb{R}^p$ represent the global shared constraints among agents. However, these global constraints are not directly observed by any agent; each agent only knows its local component $g_v(z_v, \z_{-v})$. Thus, for each agent,
\begin{align*}
    g_v(z_v, \z^g_{-v}) =& A(m_v, n_v) \, z_v +  \sum_{j \in \mathcal{G}_g} A(m_v, n_j) \, z_{j}- \mathbf{c}(m_v),
\end{align*}
where $A(m_v, n_v)$ represents the $m_v$ rows and $n_v$ columns of $A$ associated with $z_v \in \R^{n_v}$ and $\mathbf{c}(m_v)$ represents the $m_v$ rows of $\mathbf{c}$ associated with $z_v$.

The solution concept for the GNEP (\ref{equ:GNEP}) is called Generalized Nash Equilibrium (GNE):
\begin{definition}[Generalized Nash Equilibrium (GNE)]\label{def:GNE}
    $\z^\star$ is a Generalized Nash Equilibrium (GNE) if for all agent $v$,
    \begin{equation*}
    \begin{split}
        &f_v(z_v^\star,\, \z_{-v}^\star) \le f_v(z_v,\, \z_{-v}^\star), \\
        &\forall \, z_v^\star, \, z_v \in \{z_v \, | \, g_v(z_v,\, \z_{-v}^\star) = \mathbf{0}_{m_v}, \, h_v(z_v) \le \mathbf{0}_{q_v}\}.
    \end{split}
    \end{equation*}
\end{definition}
\begin{assumption}\label{ass:standard}
For every player $v$, $f_v(z_v, \z_{-v})$ and $h_v(z_v)$ are continuously differentiable, convex, and have locally Lipschitz continuous gradients with respect to $z_v$. \end{assumption}

\begin{assumption}\label{ass:KKT}
    Let $\z_{-v}^\star$be given, and suppose that $z_v^\star$ is a solution of the GNEP (\ref{equ:GNEP}), then for all $v$, $\lambda_{v}^\star \in \R^{m_v}$, $\mu_v^\star \in \R^{q_v}$ exist such that
    \begin{equation}\label{equ:kkt_gnep}
    \begin{split}
        &L_v(\z, \lambda_{v}, \mu_v): = f_v(z_v, \z_{-v}) + g_{v}^\top(z_v, \z_{-v}) \lambda_{v} + h_v^\top(z_v) \mu_v\\
        &\nabla_{z_v}  L_v(\z^\star, \lambda_{v}^\star, \mu_v^\star) = \mathbf{0}_{n_v} \\
        & g_{v}(z_v^\star,\, \z_{-v}^\star) = \mathbf{0}_{m_v}\\
        &\mathbf{0}_{q_v} \le \mu_v^\star \, \bot \, h_{v}( z_v^\star) \le \mathbf{0}_{q_v}
    \end{split},
    \end{equation}
where the last line uses the standard compact notation in optimization theory to jointly express primal feasibility, dual feasibility, and complementary slackness.
\end{assumption}
Assumption~\ref{ass:KKT} implies that, for each agent $v$, a suitable constraint qualification holds. In addition, we assume the Linear Independence Constraint Qualification (LICQ). Under the convexity assumptions in Assumption~\ref{ass:standard}, stacking the KKT conditions of all agents together with the corresponding constraint qualifications implies that $\z^\star$ is a GNE if and only if it satisfies the KKT conditions~\eqref{equ:kkt_gnep} (see Section~4.2 of \cite{facchinei-generalized-2010}).

In addition, we define the pseudo-gradient $F: \R^n \to \R^n$ as a point-to-point map, where the $v$-th block is the gradient of the cost function of agent $v$ with respect to its own decision variables, given by
\begin{align}\label{equ:pseudo_grad}
    F(\z) := \begin{bmatrix}
    \nabla_{z_1} f_{1}(z_1, \z_{-1}) \\ \nabla_{z_2} f_{2}(z_2, \z_{-2}) \\ \vdots\\ \nabla_{z_N}f_{N}(z_N, \z_{-N})
\end{bmatrix}.
\end{align}
\begin{definition}[Strongly Monotone Game]
    We say the GNEP (\ref{equ:GNEP}) is strong monotone if there exists a $ \delta>0$ such that $\forall\, \z$,  $\z' \in \R^n$,
    \begin{equation*}
        \begin{split}
        (\z - \z')^\top \left (F(\z) - F(\z') \right)  \ge  \delta \, \| \z - \z'\|^2.
        \end{split}
    \end{equation*}
\end{definition}

\subsection{Variational GNE (v-GNE)}
A Variational Inequality (VI) problem, $VI(K, F)$, is defined as:
\begin{equation}\label{equ:VI}
    \begin{split}
    &\text{Find $\z^\star \in K$ such that }\\ 
    &(\z - \mathbf{z}^\star)^\top F(\z^\star) \geq 0\qquad \forall\, \z \in K,
    \end{split}
\end{equation}
where 
\begin{equation}
    K := \left \{\z \, | \, A \, \z -\mathbf{c} = \mathbf{0}_p,  \, h_v(z_v) \le \mathbf{0}_{q_v}, \, \forall \, v \right \},
\end{equation}
is the Cartesian product of the local feasible sets and the global shared constraints. In practice, this global set is unknown to all agents. Each agent only knows a portion of it, based on its own decision variables.

In~\cite{harker-1991,facchinei-generalized-2007}, the connection between the GNEP~\eqref{equ:GNEP} and the VI problem~\eqref{equ:VI} was established for the case where all agents share global constraints as in~\eqref{equ:linear}. It was shown that solving the VI~\eqref{equ:VI} yields a solution to the GNEP~\eqref{equ:GNEP}. Such solutions are called v-GNE. It is important to note that GNEs may not be unique, and every v-GNE is contained in the solution set of the GNEP (\ref{equ:GNEP}).
\begin{assumption}\label{ass:stronglymonotone}
The GNEP~(\ref{equ:GNEP}) is strongly monotone. Moreover, $F$ is locally Lipschitz in $\z$, i.e., Lipschitz continuous on every compact set $K' \subseteq \R^n$.
\end{assumption}

Assumption~\ref{ass:stronglymonotone} guarantees a unique solution to $VI(K,F)$ and thus ensures the existence of a GNE~\cite{glynn-robinson-facchinei-2004}, even though the game itself may admit multiple equilibria. This assumption is common in distributed algorithms and arises in many engineering applications requiring distributed solvers~\cite{zhou-generalized-2005, maiorano-dynamics-2000, yi-pavel-2019, cenedese-asynchronous-2021, huang-distributed-2021, bianchi-continuous-time-2021}. Moreover, \cite{facchinei-generalized-2007} shows that a GNE of~\eqref{equ:GNEP} coincides with a solution of $VI(K,F)$ when all agents use identical multipliers for shared constraints, linking multiplier consensus to v-GNEs. A v-GNE imposes uniform shadow prices on the shared constraints, meaning the KKT multipliers are identical across all agents. This formulation implicitly assumes a coordinating authority that enforces the shared constraints and broadcasts a common shadow price. In contrast, a general (non-variational) GNE allows player-specific multipliers, so each agent faces personalized prices or tariffs on shared resources. This distinction is important in both modeling and practice, as the two formulations capture different operational trade-offs.

Our algorithm targets the general GNE set, extending applicability beyond v-GNEs. While multiplier consensus can recover a v-GNE under centralized coordination, our main result shows that even without such consensus, the proposed distributed continuous-time algorithm converges to the KKT points of the GNEP~\eqref{equ:GNEP}, and hence to a GNE. To the best of our knowledge, this is the first distributed algorithm that achieves convergence to any GNE without requiring multiplier consensus.

\subsection{Projected Dynamical System (PDS)}
We briefly review the Projected Dynamical System, a key tool in our convergence analysis.
\begin{definition}[Projected Dynamical System]\cite{debreu-social-1952}.
    Given $x \in \hat{K}$ and $v \in \R^n$, where $\hat{K} \subseteq \R^n$ a closed convex set, define the projection of the vector $v$ at $x$ with respect to $\hat{K}$ by  
    \begin{equation*}
        \Pi_{\hat{K}}(x, v) := \lim_{\epsilon \to 0} \frac{P_{\hat{K}}(x + \epsilon v) - x}{\epsilon},
    \end{equation*}
    where $P_{\hat{K}}(x) := \arg \min_{y \in \hat{K}} \| x - y\|$. Let $\hat{F}:\R^n \to \R^n$ be a given mapping. The projected dynamical system (PDS) is defined as
    \begin{equation}\label{equ:PDS}
        \dot{x} = \Pi_{\hat{K}}(x, -\hat{F}(x)).
    \end{equation}
\end{definition}
Due to the discontinuous nature of projected dynamical systems \cite{bacciotti-nonpathological-2006, cherukuri-asymptotic-2016}, the Lie derivative of a continuously differentiable function $V: \R^n \to \R$ with respect to the dynamics~(\ref{equ:PDS}) at a point $x$ is defined as
\begin{equation}\label{equ:lie_der}
\V := (\nabla_x V)^\top \Pi_{\hat{K}}(x, -\hat{F}(x)).
\end{equation}
This enables us to extend standard stability analysis and apply invariance principles.

\section{Proposed Fully Distributed Algorithm}

From the formulation~\eqref{equ:GNEP}, the distributed setup is described by two graphs: $\mathcal{G}_f=(\mathcal{V}_f,\mathcal{E}_f)$, capturing cost-function dependencies, and $\mathcal{G}_g=(\mathcal{V}_g,\mathcal{E}_g)$, capturing dependencies from shared linear constraints. Each agent knows only its own $f_v(\cdot)$, $h_v(\cdot)$, and $g_v(\cdot)$. The algorithm’s objective is to converge to a game solution while limiting information exchange over these graphs and preserving data privacy. In the worst case, both graphs are complete, with $f_v(\cdot)$ and $g_v(\cdot)$ depending on all agents’ decision variables. Building on this setup, our main contribution is Algorithm~\ref{alg:equa}.

\begin{algorithm}
\begin{algorithmic}[1]
\STATE {\bfseries Input:}Initial $z_v(0)$, $\lambda_v(0)$, $\mu_v(0) \geq 0$
\STATE $\dot{z}_v = -\nabla_{z_v} L_v(\z, \lambda_v, \mu_v)$.
\STATE $\dot{\lambda}_{v} = g_v(z_v, \z_{-v})$.
\STATE $\dot{\mu}_v^j= [h_v^j(z_v)]_{\mu_v^j}^+$, for $j \in [q_v]$.
\end{algorithmic}
\caption{Continuous Time Dynamics for Each Agent $v$}
\label{alg:equa}	
\end{algorithm}

This fully distributed method preserves the privacy of each agent’s cost function, multipliers, and constraints: agents exchange only their decision variables at each iteration. Specifically, in line~2, decision variables are communicated or made known over the cost-function graph $\mathcal{G}_f$, and in line~3 over the constraint graph $\mathcal{G}_g$. Hence, the information exchanged at each step is limited to decision variables transmitted over $\mathcal{G}_f \cup \mathcal{G}_g$. In contrast, existing v-GNE algorithms (e.g., \cite{yi-pavel-2019, cenedese-asynchronous-2021, huang-distributed-2021, bianchi-continuous-time-2021}) require exchanging not only decision variables but also Lagrangian multipliers and auxiliary variables over $\mathcal{G}_g$.

To quantify the savings, suppose $\mathcal{G}_f = \mathcal{G}_g$ and each edge transmits a vector of dimension $d$. In our algorithm, agents exchange only decision variables, whereas consensus-based methods require an additional $2|\mathcal{E}_g| \, d$ messages per fixed time for multipliers and auxiliary variables. In this special case, our approach reduces the communication cost to one-third compared to consensus-based alternatives.

\subsection{Convergence Analysis}
\begin{lemma}\label{lem:pds} Algorithm \ref{alg:equa} is an implementation of the PDS (\ref{equ:PDS}) where $\hat{F}$ is created by stacking the right-hand side of the dynamics without the projection operations, and $\hat{K} = \{(\z, \mu_v\, \lambda_v) \, | \, \z \in \R^n, \, \mu_v \ge \mathbf{0}_{q_v},\, \lambda_v \in \R^{m_v}  \text{ for all }v\}$.
\end{lemma}

Consider the set of $\lambda_v^i(t)$ evolving according to the shared constraint $g^i(\z)$, which represents the $i$-th component of the common constraints $g(\z)$. We arbitrarily select one of these as a reference state, denoted by $\lambda^i(t)$. Thus, there are $p$ reference states.

For each $\lambda_u^i(t)$ associated with the shared constraint $g^i(\mathbf{z})$ but not chosen as the reference state, we define its deviation from the reference $\lambda^i$ as $\Delta_{u}^i := \lambda_u^i - \lambda^i$. The corresponding dynamics are given by
\begin{equation}\label{equ:delta}
\dot{\Delta}_{u}^i = \dot{\lambda}_u^i - \dot{\lambda}^i = g^i(\z) - g^i(\z) = 0.
\end{equation}
 In other words, all the $\Delta_{u}^i(t)$ remain constant over time. If $u$ is selected as the reference state, $\Delta_u^i(t) = 0$. Additionally, $\Delta_v(t)$ is a stack of all  $\Delta_v^i(t)$, $i \in m_v$ with proper order. 
\begin{lemma}\label{lem:equipt}
    For any given set of $\Delta_v(0)$ for all $v$, there exists a unique equilibrium point. However, agents are not aware of this equilibrium because they do not know their respective $\Delta_v(0)$. Moreover, every equilibrium point of the Algorithm \ref{alg:equa} satisfies the KKT conditions (\ref{equ:kkt_gnep}) of the GNEP (\ref{equ:GNEP}). 
\end{lemma}

Lemma~\ref{lem:equipt} states that if a central authority has knowledge of all agents' initial values, $\lambda_v(0)$, as well as their cost functions and constraints, it can predict the equilibrium point because there is only one possible equilibrium once $\Delta_v(0)$ is set. However, individual agents cannot predict the equilibrium, as they have access to neither other agents’ $\lambda_u$ nor their cost functions and constraints. Moreover, even though the central authority can predict where the equilibrium will be, this does not guarantee that the algorithm will actually converge to that point. Given that the entire algorithm operates as a PDS, we establish the subsequent Lemma \ref{lem:monoton}.
\begin{lemma}\label{lem:monoton}
    Define $\lambda(t)$, $\Delta_v(t)$ as in (\ref{equ:delta}). Consider a positive semi-definite function $V$ defined on $\R^n \times \R^p \times \R^{q}_{\ge 0}$,
\begin{equation}\label{equ:func}
\begin{split}
    &V(\z, \lambda, \mu) \\:= &\frac{1}{2} \left [ \sum_{v = 1}^N \left ( \| z_v - z_v^\star\|^2 +  \|\mu_v - \mu_v^\star\|^2 \right ) +\| \lambda- \lambda^\star\|^2 \right ]
\end{split},
\end{equation}
where $(\z^\star, \lambda^\star, \mu_v^\star)$ is an equilibrium point as in Lemma \ref{lem:equipt}.

Then the Lie derivative of the function (\ref{equ:func}) along the trajectory $\V \leq 0$ for all $\z$, $\lambda$, and $\mu_v$. 
\end{lemma}

Using Lemma \ref{lem:monoton} and following a similar argument in Lemma 4.3 of \cite{cherukuri-asymptotic-2016}, along with Lemma \ref{lem:pds}, we derive Lemma \ref{lem:uniq}, ensuring the uniqueness and continuity of the trajectory. With this result and the fact that the Lie derivative is non-positive in Lemma \ref{lem:monoton}, we are prepared to establish the convergence of Algorithm \ref{alg:equa} in Theorem \ref{thm:main}.
\begin{lemma}\label{lem:uniq} Starting from any point $\left (\z, \lambda, \mu \right ) \in \R^n \times \R^p \times \R^{q}_{\ge 0}$, a unique trajectory $t \mapsto \gamma(t)$ to the Algorithm \ref{alg:equa} exists and remains within the set $\left(\R^n \times \R^p \times \R^{q}_{\ge 0} \right) \cap V^{-1}(\le V(\z, \lambda, \mu))$. Moreover, if a sequence of points $\{ (\z_k, \lambda_k, \mu_k)\}_{k=1}^\infty \subset\R^n \times \R^p \times \R^{q}_{\ge 0}$ converge to $(\z, \lambda, \mu)$ as $k \to \infty$, then the sequence of trajectories $\{ t \mapsto \gamma_k(t)\}_{k=1}^\infty$ starting at these points converge uniformly to the trajectory $t \mapsto \gamma(t)$ on every compact set of $[0, \infty)$.
\end{lemma}
\begin{theorem}\label{thm:main}
The set of equilibrium points of Algorithm \ref{alg:equa} is globally stable, and every trajectory converges to a GNE.
\end{theorem}

\begin{remark}
If all agents initialize with identical multipliers (e.g., set to zero by default or centrally), they can converge to the same v-GNE without exchanging multiplier information. We see that our approach can recover the v-GNE as existing schemes, as a special case.
\end{remark}

\subsection{Discretization for Discrete-Time Algorithms}

Although the continuous-time algorithm has elegant theoretical properties, it must be discretized for implementation. Discretization, however, introduces approximation errors and often requires additional conditions to preserve the behavior of the continuous dynamics. Accordingly, we develop two discretization schemes for Algorithm~\ref{alg:equa}, each with its own assumptions determined by the chosen technique. In addition to the standard Assumptions~\ref{ass:standard}--\ref{ass:stronglymonotone}, each scheme requires at least one additional assumption to guarantee convergence.
\begin{assumption}[Additional Assumptions]\label{ass:discrete}
The discretized algorithms require one or more of the following assumptions:
\begin{enumerate}
    \item The pseudo-gradient in~(\ref{equ:pseudo_grad}) is globally Lipschitz with constant $L_f$. 
    \item For every agent $v$ and any $z_v, z_v' \in \mathbb{R}^{n_v}$,
    \begin{equation*}
        \|h_v(z_v) - h_v(z_v')\|^2 \le L_v \|z_v - z_v'\|^2.
    \end{equation*}
    \item The individual constraint set $Z_v := \{z_v \,|\, h_v(z_v) \le \mathbf{0}\}$ is compact for all agents.
\end{enumerate}
\end{assumption}

Based on \cite{ebrahimi-robust-2019}, Algorithm~\ref{alg:disc_equa} provides a discrete version of the continuous algorithm using a standard diminishing step size. Specifically, we choose a step size $\alpha^k$, where $k$ denotes the iteration index, satisfying:
\begin{equation}\label{equ:disminishing_step_size}
    \begin{split}
        \sum_{k=1}^\infty \alpha^k = \infty, \,  \sum_{k=1}^\infty (\alpha^k)^2 = \Gamma < \infty.
    \end{split}
\end{equation}
It is important to note that this step size is independent of agents' private data and can therefore be chosen without accessing such information. To establish convergence, we require a deterministic version of the almost supermartingale convergence lemma from \cite{ROBBINS1971233}.

\begin{algorithm}
\begin{algorithmic}[1]
\STATE {\bfseries Input:}Initial $z_v^0$, $\lambda_v^0$, $\mu_v^0 \geq 0$, $\alpha^k$.
\FOR{$k=1, 2, \dots, K$}
\STATE Receive the decision variables $\z_{-v}$ from other agents through the cost-function graph $\mathcal{G}_f$ and the shared constraint graph $\mathcal{G}_g$.
\STATE $z_v^{+} \leftarrow z_v-\alpha^k \cdot \nabla_{z_v} L_v(\z, \lambda_v, \mu_v)$.
\STATE $\lambda_{v} \leftarrow \lambda_v + \alpha^k \cdot g_v(z_v, \z_{-v})$.
\STATE $\mu_v^j \leftarrow P_{\ge0} \left[ \mu_v^j + \alpha^k \cdot h_v^j(z_v) \right ]$, for $j \in [q_v]$.
\STATE $z_v \leftarrow z_v^+$.
\ENDFOR
\end{algorithmic}
\caption{Forward Discretization for Each Agent $v$}
\label{alg:disc_equa}	
\end{algorithm}

\begin{theorem}\label{thm:discrete}
With a diminishing step size as in (\ref{equ:disminishing_step_size}), and under the additional Assumptions~\ref{ass:discrete} (1) and (2), the sequence ${\z^k}$ generated by Algorithm~\ref{alg:disc_equa} converges to a fixed point $\z^\star$, which is a GNE.
\end{theorem}

Theorem~\ref{thm:discrete} shows that a diminishing step size yields a valid discretization of Algorithm~\ref{alg:equa}, ensuring convergence under additional assumptions on the GNEP. This also suggests that other discretizations, such as Euler with a constant step size, may be suitable for GNEPs with different properties. Beyond forward discretization, backward (resolvent) or hybrid schemes can also be used. Algorithm~\ref{alg:disc_equa_os} illustrates a forward--backward hybrid scheme (similar in spirit to Runge--Kutta methods), in which the multiplier step size $\alpha$ must be chosen globally, i.e., all agents share the same $\alpha$ even if their initial multipliers differ.

\begin{theorem}\label{thm:alg_disc_os_converge}
Define the inverse learning-rate matrices as
\begin{equation*}
\tau := \text{diag}([\tau_1^{-1}, \dots, \tau_N^{-1}]), 
\quad 
\sigma := \alpha^{-1} \mathbf{I}_{p \times p},
\end{equation*}
where $\mathbf{I}_{p \times p}$ is the identity matrix of size $p$.
If
\begin{equation*}
\min \{ \lambda_{\min}(\tau), \, \lambda_{\min}(\sigma) \} > \tfrac{L_f}{2\delta} + \|A\|_2,
\end{equation*}
where $\lambda_{\min}(\cdot)$ denotes the smallest eigenvalue, and Assumptions~\ref{ass:discrete} (1) and (3) hold, then Algorithm~\ref{alg:disc_equa_os} converges to a fixed point corresponding to a GNE of the GNEP~\eqref{equ:GNEP}.
\end{theorem}

\begin{algorithm}
\begin{algorithmic}[1]
\STATE \textbf{Input:}Initial $z_v^0$, $\lambda_v^0$, $\tau_v$, $\sigma$.
\FOR{$k=1, 2, \dots, K$}
\STATE Receive the decision variables $\z_{-v}$ from other agents through the cost-function graph $\mathcal{G}_f$ and the shared constraint graph $\mathcal{G}_g$.
\STATE \begin{flalign*}
z_v^+ \leftarrow P_{Z_v} \Big\{ z_v - \tau_v \cdot \big[ 
    & \nabla_{z_v} f_v(z_v, z_{-v}) \\
    & + \lambda_v^\top \nabla_{z_v} g_v(z_v, \z_{-v}) \big] \Big\}. &&
\end{flalign*}
\STATE Receive the updated decision variables $\z_{-v}^+$ from other agents through the shared-constraint graph $\mathcal{G}_g$.
\STATE \begin{flalign*}
\lambda_{v} \leftarrow \lambda_v + \alpha \cdot \big[ 
    & 2 \cdot g_v(z_v^+, \z_{-v}^+) - g_v(z_v, \z_{-v}) \\
    & + 2 \cdot \mathbf{c}(m_v) \big]. &&
\end{flalign*}
\STATE $z_v \leftarrow z_v^+$.
\ENDFOR
\end{algorithmic}
\caption{Forward-Backward for Each Agent $v$}
\label{alg:disc_equa_os}	
\end{algorithm}

\section{Multi-robots Placement Problem}

We apply our fully distributed algorithm to a multi-robot placement problem. Multiple agents (robots) act as simple integrators and collaborate to complete shared tasks. Each task $j$ has a center location $c_j \in \mathbb{R}^2$ and an associated set of robots $N_j$ with size $n_j$. A robot $v$ may be assigned to multiple tasks, each with a different center.

This setup induces global shared constraints: for each task $j$, the positions of all assigned robots must collectively center at $c_j$. Mathematically, this is represented as $g_v^j(z_v, \z_{-v}) = \sum_{v \in N_j} z_v - n_j \cdot c_j=0$, where $z_v = (x_v, y_v)$ denotes the placement of robot $v$. Let $M_v$ be the set of tasks involving robot $v$. Then $g_v(z_v, \z_{-v})$ is the collection of all constraints $g_v^j(z_v, \z_{-v})$ for $j \in M_v$. Importantly, while a global constraint set $g(\z)$ exists, each robot only has access to its local constraints $g_v(z_v, \z_{-v})$, not the full global set.

In addition to task locations, each robot $v$ has an anchor point $A_v$, representing its charging or control station. Each robot must remain within a range centered at $A_v$, where both the radius and location depend on the robot’s capabilities. Formally, this local constraint is $\|z_v - A_v\| \le r_v$, which depends only on robot $v$ and is independent of other robots’ positions.

\begin{figure}[H]
\centering
\includegraphics[width=0.8\columnwidth]{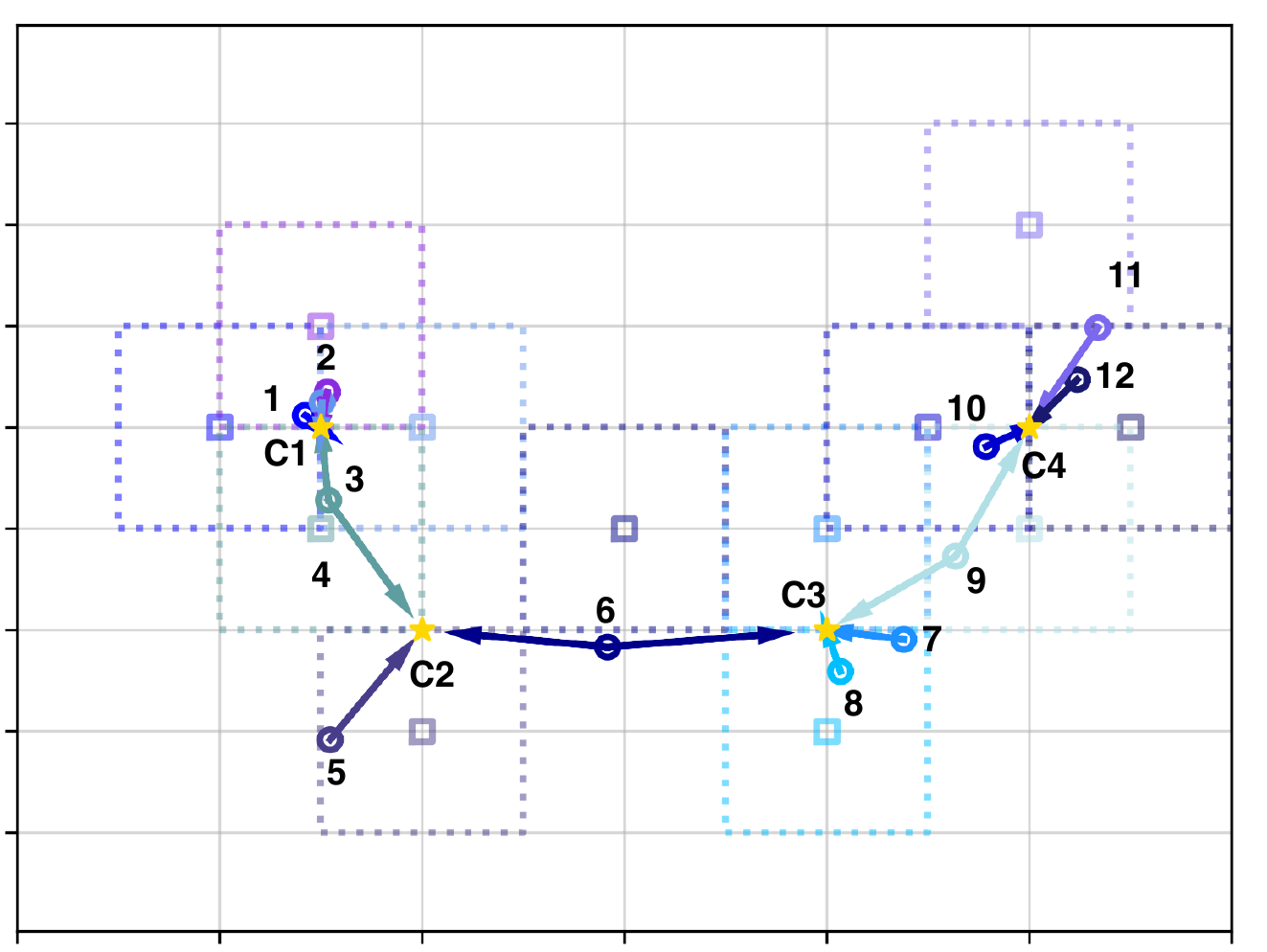} 
    \caption{Multi-robot placement with 12 robots and 4 tasks. Circles denote robots $v$, rectangles their anchor points $A_v$, and dotted boxes local constraint regions. Yellow stars $c_j$ mark task locations. Arrows indicate assignments ($v \in N_j$, $j \in M_v$), forming a bipartite robot–task structure.}
    \label{fig:setup}
\end{figure}

\begin{figure}[H]
\centering
\includegraphics[width=0.78\columnwidth]{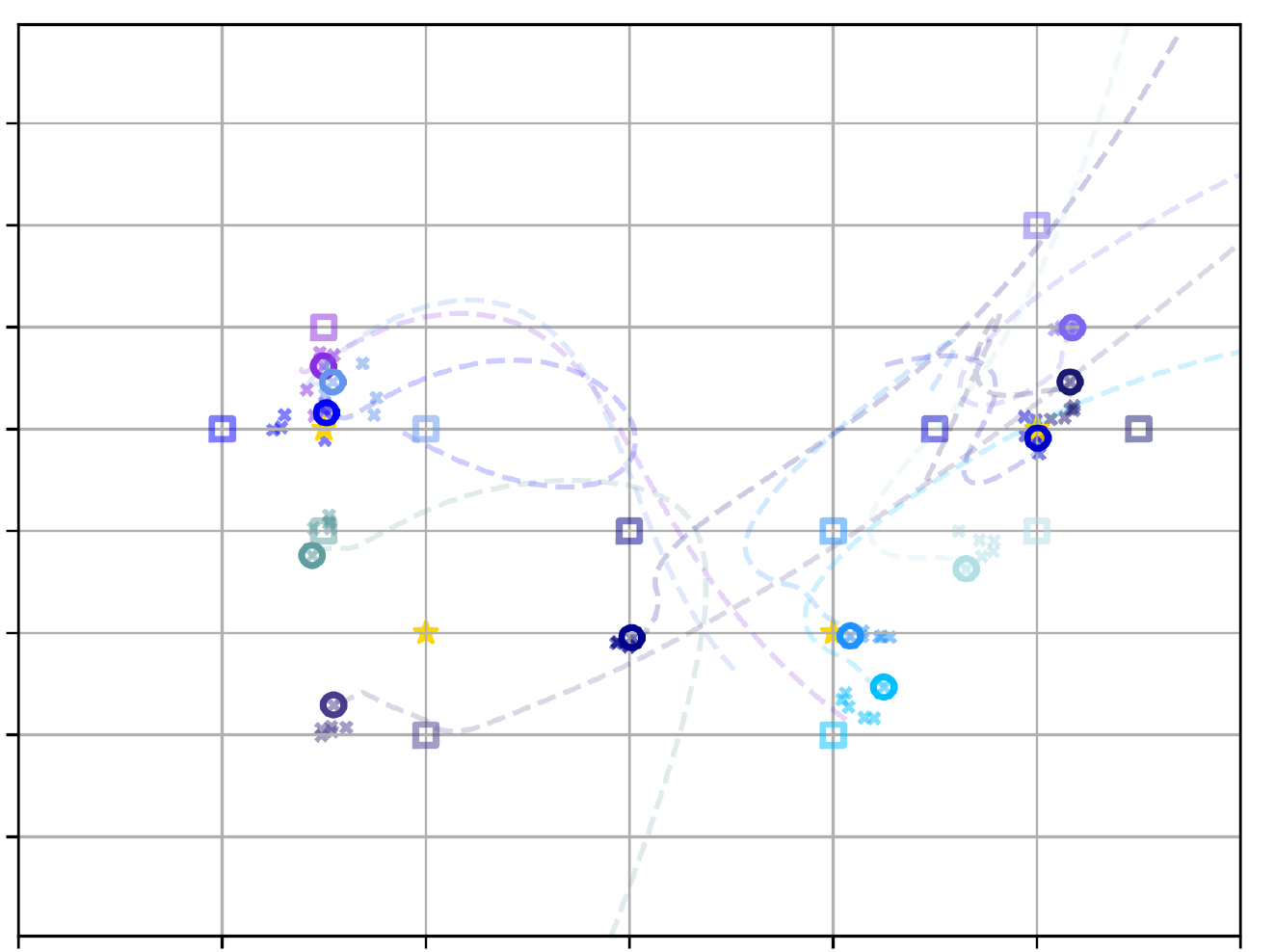}
    \caption{Converging trajectories from six random initializations (seeds 0--5). ``x'' marks the corresponding convergence points, all of which are GNEs.}
    \label{fig:trajectory}
\end{figure}

For the objective function, each robot balances two goals: staying close to its anchor point and staying close to other robots $u \in N_j$ for tasks $j \in M_v$ to enable coordination. To capture this trade-off, each agent’s cost function includes a private penalty parameter $\rho_v$. Specifically, the cost function is defined as:
\begin{equation}
    f_v(z_v, \z_{-v}) = \| z_v - A_v\|^2 + \rho_v \sum_{u \in N_j, j \in M_v} \| z_v - z_u\|^2.
\end{equation}
This leads to the following GNEP with convex individual constraints and linear shared constraints:
\begin{equation*}
    \begin{matrix*}[l]
        \min_{z_v} & f_v(z_v, \z_{-v}) \\
        \text{subject to} & \| z_v - A_v\| \le r_v\\
        &g_v(z_v,\, \z_{-v}) = 0
    \end{matrix*}.
\end{equation*}
Figure~\ref{fig:setup} illustrates a setup with 12 robots and 4 tasks. Our algorithm is fully distributed: each robot keeps its constraint and cost information locally and does not exchange multipliers. Since GNEs depend on initialization, Figure~\ref{fig:trajectory} shows different equilibria reached by the continuous algorithm~\ref{alg:equa} under varying initial conditions. This non-uniqueness highlights both the difficulty of computing such equilibria, even with centralized solvers, and the contribution of our method.

\begin{figure}[H]
\centering
\vspace{2mm}
\includegraphics[width=0.8\columnwidth]{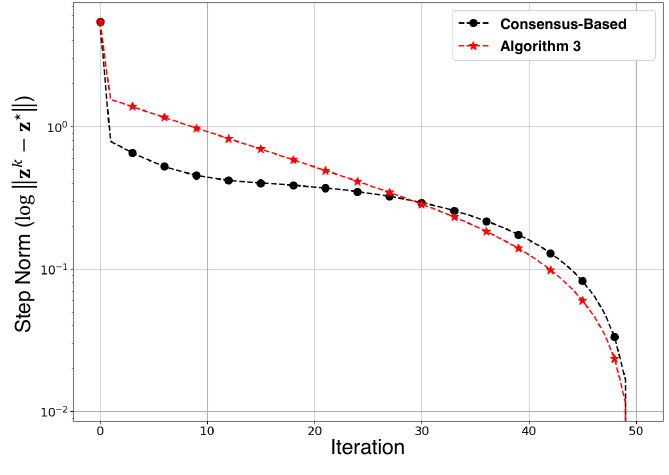}
    \caption{
Comparison of Algorithm~\ref{alg:disc_equa_os} (forward–backward discretization) with consensus-based approaches from~\cite{yi-pavel-2019}. Both are initialized at zero with learning rate $0.01$, converging to a v-GNE.
    }
\label{fig:compare_consensus_trajectory}
\end{figure}

Figure~\ref{fig:compare_consensus_trajectory} compares the convergence of our forward--backward discretization algorithm~\ref{alg:disc_equa_os} with a consensus-based method adapted from~\cite{yi-pavel-2019}. Both show similar convergence experimentally, but our method requires less communication per iteration: since $\mathcal{G}_f$ coincides with $\mathcal{G}_g$, information exchange is reduced by two-thirds per iteration.


\section{Conclusion}
We proposed a fully distributed continuous-time algorithm for solving GNEPs without multiplier consensus. By eliminating multiplier exchange, the method reduces communication overhead while converging to the broader set of GNEs; v-GNEs can still be recovered by identical multiplier initialization. Discrete-time variants were also introduced and validated on a multi-robot placement problem with shared constraints. Future work includes extending the analysis to GNEPs with global inequality constraints, where formal convergence guarantees remain open, and studying how sparsity and graph structure affect communication efficiency, in particular whether $\mathcal{G}_g$ can be chosen sparser than $\mathcal{G}_f$ and what the sparsest feasible communication graph is, especially in dynamic networks where agents may join or leave.





\bibliography{IEEEexample}
\bibliographystyle{IEEEtran}

\end{document}